\documentclass[11pt]{article}

\usepackage[utf8]{inputenc}
\usepackage[T1]{fontenc}
\usepackage[margin=1in]{geometry}
\usepackage{mathptmx}
\usepackage{amsmath,amssymb}
\usepackage{graphicx}
\usepackage{booktabs}
\usepackage{array}
\usepackage{tabularx}
\usepackage{longtable}
\usepackage{float}
\usepackage{needspace}
\usepackage[font=small,labelfont=bf]{caption}
\usepackage[numbers,sort&compress]{natbib}
\usepackage[colorlinks=true,linkcolor=blue!50!black,citecolor=blue!50!black,urlcolor=blue!50!black]{hyperref}
\usepackage{enumitem}
\usepackage{xcolor}
\usepackage{seqsplit}
\newcommand{\hashtt}[1]{\texttt{\seqsplit{#1}}}

\title{\textbf{CareGraph: An Auditable Hybrid AI Framework for\\
Evidence-Grounded Personalized Longitudinal Health Intelligence}}

\author{
Pratik Ghawate \quad Tanvi Patil \\[4pt]
\small California, United States
}
\date{}

\begin{document}

\maketitle

\begin{center}
\small\itshape
Controlled synthetic research evaluation of a personalized longitudinal health-intelligence framework for integrating multi-source health data; intended for information organization and clinician-discussion support.
\end{center}

\vspace{0.5em}

\begin{abstract}
\noindent
Health information already exists across laboratory portals, electronic health records, medication lists, symptoms, lifestyle reports, and wearable devices, yet these sources rarely form a coherent and traceable explanation for an individual. CareGraph is proposed as an auditable hybrid AI framework and extensible backbone for evidence-grounded personalized longitudinal health intelligence. It transforms heterogeneous patient-specific records into prioritized trends, missing-context indicators, bounded next steps, clinician-discussion questions, and traceable explanations. Personalization denotes organization, prioritization, and interpretation of an individual's available evidence; it does not denote diagnosis, prognosis, treatment selection, or autonomous clinical decision-making. The architecture separates deterministic data-sufficiency and trend analysis, scenario-aware missing-context detection, patient-state assembly, provenance-oriented graph construction, schema-constrained language-model synthesis, evidence validation, safety transformation, and release gating. Independently seeded synthetic development, validation, and held-out cohorts contained 400 patients each. A frozen ordinary-least-squares trend rule with an explicit sufficiency gate achieved held-out accuracy 0.827, macro-F1 0.837 (95\% CI 0.819--0.854), and insufficient-data F1 0.974. Schema-aligned missing-context detection achieved strict held-out micro-F1 0.815 versus 0.318 for the legacy detector, and safety ruleset v1.2 achieved precision 1.000, recall 0.950, and F1 0.974 on a held-out authored benchmark. A graph-required 80-patient engineering audit produced 79 AI syntheses and 78 presentation outputs with no fallback; one output was blocked and one failed closed after an invalid evidence key. In a matched 56-patient comparison with an isolated monolithic GPT-5.6 baseline, CareGraph was faster (40.15 versus 49.62~s), shorter (661 versus 1{,}163 words), and more aligned with predefined longitudinal targets in exploratory lexical checks, while the monolithic baseline used fewer total tokens and cited more direct raw-record evidence. The graph audit verified provenance integrity and deterministic retrieval; incremental graph contribution to generated content is reserved for a dedicated paired evaluation. These results establish CareGraph as a technically validated foundation for a new class of patient-facing and clinician-facing personalized health-intelligence systems within a controlled synthetic setting.
\end{abstract}


\section{Introduction}

Longitudinal health information is distributed across laboratory portals, electronic health records, medication lists, symptoms, vital signs, lifestyle reports, family and medical history, and wearable devices. The central systems problem is therefore not simply data availability; it is fragmentation. Individuals and clinicians often see isolated measurements without a coherent account of what changed over time, what information is missing or contradictory, what deserves discussion, and where each statement came from. A useful health-intelligence system must provide that synthesis without hiding calculation, uncertainty, provenance, or release controls.

Large language models (LLMs) demonstrate substantial medical question-answering capability \citep{singhal2023, singhal2025}, but fluency can obscure unsupported claims, omissions, and overconfident recommendations \citep{asgari2025}. CareGraph addresses this limitation by treating personalized health intelligence as a systems problem rather than a one-pass prompting problem. Deterministic and rule-based stages calculate and qualify patient-specific evidence; an optional LLM converts only authorized evidence into a constrained schema; and a separate safety stage validates references and controls release. The objective is not autonomous clinical decision-making, but an intelligible layer that helps a person or clinician understand longitudinal patterns and prepare better questions.

CareGraph is designed as an extensible health-intelligence backbone rather than a closed application. New laboratory, EHR, medication, symptom, lifestyle, or wearable adapters can be normalized into the same patient state and authorized evidence registry, while downstream trend logic, provenance, synthesis, and release contracts remain stable. In the envisioned use, heterogeneous health records can be connected to the framework and transformed into a concise explanation of what changed, what context is absent or conflicting, which issues merit discussion, and how every generated statement traces back to source data. The current study evaluates the technical foundation for that broader system vision.

The evaluation program progresses from component validity to end-to-end system behavior. Independently seeded development, validation, and held-out cohorts assess frozen trend and missing-context methods; a separate authored benchmark evaluates safety rules; a 400-patient audit verifies graph provenance; an 80-patient engineering batch measures operational reliability and release behavior; and a matched 56-patient comparison characterizes the trade-off between CareGraph and an isolated monolithic baseline. Together, these experiments test whether the proposed backbone can turn fragmented longitudinal records into focused, inspectable, and safety-governed personalized health intelligence.

The paper makes seven contributions:

\begin{itemize}[leftmargin=1.4em]
\item An extensible hybrid AI framework that separates multi-source data integration, deterministic evidence generation, graph-based provenance, constrained synthesis, safety review, and presentation gating.
\item A frozen trend protocol with independent development, validation, and held-out cohorts, an explicit data-sufficiency gate, and class-, scenario-, and biomarker-level reporting.
\item A schema-aligned missing-context detector with exact versioned canonicalisation, backward-compatible legacy output, and an audit of generator-ground-truth limitations.
\item A transparent safety ruleset evaluated on separate authored development and held-out adversarial examples, including retained false-negative analysis.
\item A 400-patient graph retrieval and provenance audit that separates structural lineage verification from unproven graph-specific content benefit.
\item An 80-patient end-to-end synthesis audit reporting completion, evidence validity, latency, prompt drift, safety interventions, one fail-closed invalid-key error, and one apparent false-positive safety block.
\item A matched 56-patient final-prompt comparison with an isolated monolithic baseline, reporting paired generation efficiency, output breadth, evidence use, and exploratory longitudinal-coverage checks.
\end{itemize}

\section{Related Work}

\subsection{Longitudinal Electronic Health Records}

Longitudinal electronic health records are irregular, heterogeneous, sparse, and institution dependent. Large-scale predictive systems can learn from broadly encoded EHR data \citep{rajkomar2018}, while temporal EHR reviews emphasise missingness, varying intervals, and inconsistent representations \citep{xie2022}. CareGraph does not train a clinical prediction model; it focuses on transparent patient-level evidence preparation and bounded synthesis.

\subsection{Medical Large Language Models}

Medical LLM studies show that scale and instruction design can improve medical question answering \citep{singhal2023, singhal2025}. However, medical summarisation can contain factual inconsistencies, hallucinations, and unsupported implications \citep{asgari2025}. These risks motivate schema constraints, evidence identifiers, and post-generation safety checks, while not eliminating the need for semantic and expert evaluation.

\subsection{Biomedical Knowledge Graphs}

Biomedical knowledge graphs can integrate heterogeneous biomedical entities and relations \citep{chandak2023}, and graph representation learning can support biomedical inference \citep{li2022}. CareGraph uses a much narrower patient-scoped graph: typed nodes, typed edges, source-row provenance, and deterministic evidence ranking. It does not implement graph neural networks, embeddings, causal inference, or graph-based clinical reasoning.

\subsection{Synthetic Data and Clinical AI Reporting}

Synthetic health data support software development without directly exposing protected health information, with Synthea providing a prominent example \citep{walonoski2018}. Synthetic benchmarks nevertheless inherit generator assumptions and may reward rules aligned to generator schemas. Reporting guidance for AI interventions and early-stage clinical evaluation emphasises versioning, workflow boundaries, human interaction, and failure analysis \citep{liu2020, riveracruz2020, vasey2022}. The present study is a controlled synthetic component evaluation, not a clinical trial or workflow-effectiveness study.

\section{Problem Definition and Research Questions}

For patient $p$, let the heterogeneous longitudinal record be $D_p$, drawn from profile, encounters, laboratories, medications, symptoms, vitals, lifestyle, family history, medical history, and wearables. CareGraph maps these sources into a patient state and produces a structured personalized health-intelligence output $Y_p$ containing a summary, prioritized findings, bounded next steps, clinician questions, uncertainty, and evidence references. For every generated finding or next step $z$, the structural evidence constraint is:
\[
\text{evidence}(z) \neq \emptyset \quad \text{and} \quad \text{evidence}(z) \subseteq K_p,
\]
where $K_p$ is the authorised patient evidence registry. Existing-key validity is necessary but not sufficient: semantic support additionally requires the text to accurately represent the cited evidence.

The revised study asks:

\begin{itemize}[leftmargin=2.6em]
\item[\textbf{RQ1:}] What held-out performance does the frozen trend method achieve after separating data sufficiency from directional classification?
\item[\textbf{RQ2:}] How much does schema-aligned missing-context detection improve over the legacy detector on an independently seeded held-out cohort?
\item[\textbf{RQ3:}] How accurately does the revised transparent safety ruleset detect held-out adversarial language?
\item[\textbf{RQ4:}] What structural and provenance properties can be verified for deterministic graph retrieval across the 400-patient source snapshot?
\item[\textbf{RQ5:}] What operational completion, evidence-governance, prompt-drift, and safety failure modes occur in an 80-patient graph-required engineering batch?
\item[\textbf{RQ6:}] How do final-prompt CareGraph and an isolated monolithic baseline differ on matched generation latency, token use, output length, evidence references, and exploratory longitudinal-coverage indicators?
\end{itemize}

\section{System Architecture}

\begin{figure}[H]
\centering
\includegraphics[width=\textwidth]{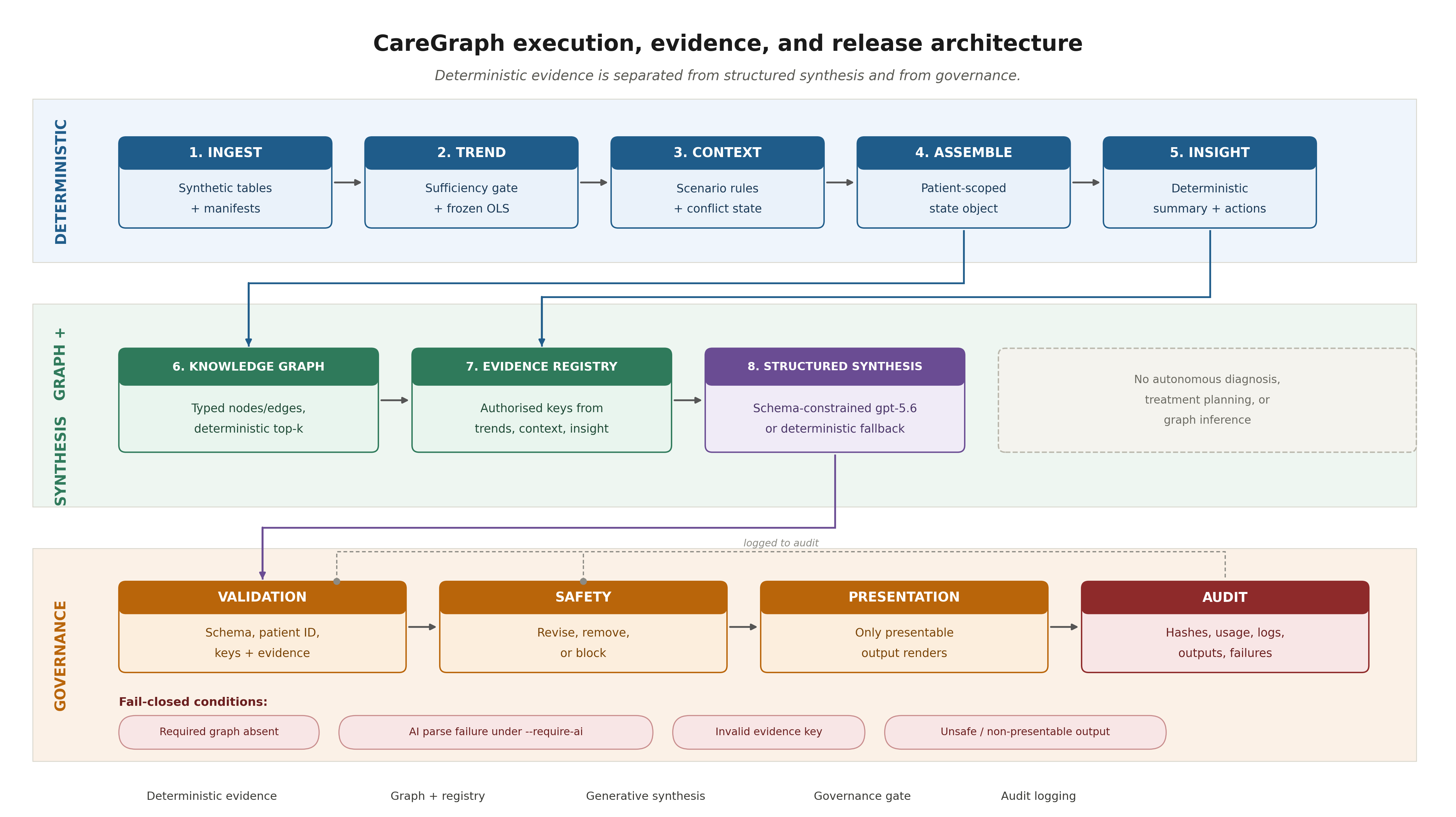}
\caption{CareGraph execution, evidence, and release architecture. Deterministic evidence is separated from structured synthesis and from governance. Invalid evidence, missing required graph/AI artefacts, and non-presentable safety outcomes fail closed.}
\label{fig:architecture}
\end{figure}

CareGraph follows four design principles: deterministic evidence before generative language; explicit representation of missing and conflicting context; evidence-keyed generation; and separation of generation from release. The architecture is intentionally source-extensible. Data adapters populate a common patient state; deterministic agents compute sufficiency, trends, and missing context; a patient-scoped graph preserves relationships and source-row lineage; an authorized registry decouples upstream integration from downstream language generation; and safety and presentation gates prevent unsupported or non-presentable outputs from silently reaching users. The implemented runtime sequence is shown in Fig.~\ref{fig:architecture}.

The architecture is governed by a patient-scoped evidence contract. Let $K_p$ denote the authorised registry formed from frozen trend outputs, missing-context states, assembled patient evidence, deterministic insight items, and retrieved graph evidence. Every generated finding and action must cite at least one key, and every cited key must belong to $K_p$.
\[
\forall\, z \in F_p \cup A_p:\quad \text{evidence}(z) \neq \emptyset \ \wedge\ \text{evidence}(z) \subseteq K_p.
\]

Generation and release are separate state transitions. A synthesis can exist without being releasable; presentation is permitted only when the schema, identity, evidence, safety, and artefact contracts all pass.
\begin{align*}
\text{Release}(Y_p) = \ & \text{schema-valid} \wedge \text{patient-match} \wedge \text{key-valid}\ \wedge {} \\
& \text{evidence-complete} \wedge \text{safety-presentable} \wedge \text{artefact-complete}.
\end{align*}

\begin{table}[H]
\centering
\caption*{\textbf{Table 1.} Component Inventory}
\label{tab:component-inventory}
\footnotesize
\begin{tabularx}{\textwidth}{@{}l l X X X@{}}
\toprule
\textbf{Component} & \textbf{Logic} & \textbf{Input} & \textbf{Output} & \textbf{Boundary} \\
\midrule
Trend Agent & Deterministic numerical & Labs + scenario inventory & Trend class + sufficiency state & No clinical significance \\[3pt]
Missing-Context Agent & Rule-based & Normalised patient domains + scenario & Canonical missing-context items & Scenario-conditioned benchmark \\[3pt]
Conflict Resolver & Rule-based & Assembled record & Conflict state & Narrow family-history scope \\[3pt]
Knowledge Graph & Deterministic transform & Patient state + upstream evidence & Nodes, edges, ranked evidence & No graph reasoning \\[3pt]
Personalised Insight & Deterministic transform & Patient state + trends + context & Structured insight & Heuristic confidence \\[3pt]
AI Synthesis & External LLM & Authorised evidence registry & Schema-constrained synthesis & Optional generative layer; no autonomous diagnosis or treatment \\[3pt]
Safety Agent & Rule-based + validation & Generated or deterministic output & Approve, revise, or block & Surface-pattern limitations \\[3pt]
Presentation Agent & Deterministic transform & Approved output & Readable artefact & Cannot bypass block \\
\bottomrule
\end{tabularx}
\end{table}

\textbf{Algorithm 1.} CareGraph Patient Inference and Release Control
\vspace{-0.3em}
\begin{longtable}{@{}p{0.06\textwidth} p{0.44\textwidth} p{0.42\textwidth}@{}}
\toprule
\textbf{Step} & \textbf{Operation} & \textbf{Fail-closed condition or recorded state} \\
\midrule
\endfirsthead
\toprule
\textbf{Step} & \textbf{Operation} & \textbf{Fail-closed condition or recorded state} \\
\midrule
\endhead
\bottomrule
\endfoot
1 & Load the versioned patient record, manifests, prompt/configuration hashes, and required runtime flags. & Missing source, malformed record, or manifest/hash mismatch. \\[4pt]
2 & Apply the sufficiency gate, classify trends, detect scenario-conditioned missing context, and resolve supported conflicts. & Emit reason-coded insufficient state; do not silently coerce unusable measurements. \\[4pt]
3 & Assemble the patient state, construct typed graph nodes/edges, and deterministically rank top-$k$ evidence. & Abort when graph is required but unavailable; retain graph provenance and retrieval metadata. \\[4pt]
4 & Construct $K_p$ and the deterministic insight payload; freeze model, schema, prompt hash, and input hash. & Reject unknown evidence types or an empty authorised registry for required synthesis. \\[4pt]
5 & Generate schema-constrained synthesis, or use deterministic fallback only when fallback is permitted. & Raise on API/parse failure under \texttt{-{}-require-ai}; record fallback reason otherwise. \\[4pt]
6 & Validate patient identity, schema, evidence-key membership, and evidence completeness for every finding and action. & Any invalid key or unsupported required item stops safety and presentation. \\[4pt]
7 & Apply safety rules, record violations and revisions, remove unsafe list items, and compute presentability. & Blocked or empty-safe-content output cannot reach presentation. \\[4pt]
8 & Render audience-specific artefacts and archive logs, usage, hashes, intermediate outputs, and failure metadata. & Missing required artefact marks the run incomplete. \\
\end{longtable}

Algorithm 1 makes the central design distinction explicit: generation is probabilistic, but admissible evidence, schema validation, failure handling, safety transformation, and release are machine-checkable.

\section{Data and Experimental Protocol}

\subsection{Independently Seeded Synthetic Cohorts}

The revised component study uses dataset version \hashtt{caregraph-synthetic-reviewer-response-v1.0}. Development, validation, and replacement held-out cohorts were generated independently with 50 patients in each of eight scenarios. Each cohort contains 400 patients, three annual encounters per patient, and 1{,}700 expected patient--laboratory trend labels. The nominal test seed 20260802 was conservatively excluded before analysis because sample records were printed to the audit console; replacement seed 20261802 was generated quietly and used once after code and configuration freeze.

\needspace{8\baselineskip}\begin{center}\small\textbf{Table 2.} Evaluation Cohorts and Roles\end{center}\label{tab:cohorts}
\vspace{-0.3em}
\begin{longtable}{@{}p{0.16\textwidth} p{0.11\textwidth} p{0.09\textwidth} p{0.10\textwidth} p{0.13\textwidth} p{0.19\textwidth}@{}}
\toprule
\textbf{Cohort} & \textbf{Seed} & \textbf{Patients} & \textbf{Trend labels} & \textbf{Missing/safety labels} & \textbf{Role} \\
\midrule
\endfirsthead
\toprule
\textbf{Cohort} & \textbf{Seed} & \textbf{Patients} & \textbf{Trend labels} & \textbf{Missing/safety labels} & \textbf{Role} \\
\midrule
\endhead
\bottomrule
\endfoot
Development & 20260731 & 400 & 1{,}700 & 567 & Method/rule development \\[3pt]
Validation & 20260801 & 400 & 1{,}700 & 587 & Selection and freeze verification \\[3pt]
Held-out test & 20261802 & 400 & 1{,}700 & 564 & One-shot final component evaluation \\[3pt]
Excluded nominal test & 20260802 & 400 & 1{,}700 & --- & Not analysed; preview exposure \\[3pt]
Safety development & Authored split & --- & --- & 49 examples & Ruleset development \\[3pt]
Safety held-out & Locked authored split & --- & --- & 39 examples & One-shot safety evaluation \\
\end{longtable}

The eight scenarios are metabolic risk, cardiovascular risk, kidney safety, liver-enzyme trend, anaemia/fatigue, thyroid/endocrine pattern, medication safety/side-effect context, and inflammation/autoimmune signal. Scenario is provided directly to the runtime inventory and missing-context rules. This explicit scenario conditioning is a benchmark limitation and potential leakage channel, not evidence of open-world generalisation.

\subsection{Reproducibility and Freeze Protocol}

The experiment version is \hashtt{caregraph-reviewer-response-v1.0}, prepared on branch \hashtt{research/reviewer-response} from starting commit \hashtt{cce87ffabae173525f82dab83fdb7cf1db93a1bf}. The initial worktree was dirty; pre-existing changes were preserved in a scoped backup with SHA-256 \hashtt{d7c8450f...7465ac3}. Source and configuration hashes were frozen before held-out evaluation. Runtime-versus-evaluator checks matched all 1{,}700 validation trend rows and all 761 validation missing-context labels.

The validation report passed all 59 required-artifact checks, parsed 150 CSV and 95 JSON files, verified source hashes and cohort manifests, retained failed commands and the single safety false negative, and confirmed the integrity of the frozen component-evaluation package. The later 80-patient operational audit and 56-patient monolithic comparison were analysed under separate manifests and prompt hashes. The audited environment used Python 3.11.3 on macOS 15.7.4 arm64 with NumPy 1.26.0, pandas 2.1.1, SciPy 1.11.3, and scikit-learn 1.4.1.post1.

\subsection{Statistical Analysis}

Classification metrics use standard one-vs-rest definitions. Trend results include accuracy, balanced accuracy, macro-F1, weighted F1, class support, scenario metrics, and biomarker metrics. Missing-context micro metrics are calculated over unique patient--scenario--canonical-field tuples. Patient-level set-F1 is paired across detectors. Safety metrics include sensitivity, specificity, false-negative rate, and false-positive rate.
\[
\text{Precision} = \frac{TP}{TP+FP}, \qquad \text{Recall} = \frac{TP}{TP+FN}, \qquad F_1 = \frac{2PR}{P+R}.
\]

Percentile 95\% confidence intervals use 2{,}000 patient-cluster bootstrap resamples for trend and missing-context metrics. The supported post-freeze component tests are a two-sided Wilcoxon signed-rank test for patient-level missing-context F1 and an exact McNemar/binomial test for safety correctness, with Holm family-wise correction across those two tests. For the matched 56-patient generation comparison, paired differences were evaluated exploratorily with two-sided Wilcoxon signed-rank tests, paired rank-biserial effect sizes, and Holm correction across nine reported generation metrics. These tests quantify operational differences, not clinical quality.

\section{Component Methods}

\subsection{Trend Agent}

The historical endpoint-change rule is retained as a baseline. The revised agent first applies an explicit sufficiency gate. A series is sufficient only when it contains at least three valid measurements, spans at least 365 days, uses one comparable unit, has parseable dates and values, contains no unresolved same-date value conflict, has no missing measurement under the frozen maximum-missingness setting, and has a nonzero scale. Failed gates emit explicit reasons and map to \texttt{insufficient\_data} only for benchmark scoring.

For a sufficient series with values $x_i$ observed at elapsed days $t_i$, the selected OLS score is the fitted slope $\beta_1$ multiplied by the observed span and normalised by the magnitude of the first value:
\[
s_{OLS} = \frac{\beta_1 (t_n - t_1)}{\max(|x_1|, \varepsilon)}.
\]
The label is increasing when $s_{OLS} > \tau$, decreasing when $s_{OLS} < -\tau$, and stable otherwise. Candidate endpoint, OLS, Theil--Sen, median-window, and transparent ensemble methods were compared using development and validation only. Macro-F1 was primary; balanced accuracy, overall accuracy, and minimum class F1 were secondary. OLS with $\tau = 0.08$ was frozen in a single active configuration. The legacy same-cohort 0.09 observation is retained only as historical context and is not a runtime setting.

\subsection{Missing-Context Agent}

Version 2 corrects source-schema mappings, uses a reviewed exact canonical map without fuzzy matching, emits obligation and context-state fields, and preserves backward-compatible legacy output. Scenario-specific rules operate over laboratories, vital signs, lifestyle, family history, medical history, medications, symptoms, and wearables. Exact canonical matching is used for the strict benchmark; the frozen alias result is numerically identical and is not described as human-adjudicated semantic performance.

The ground-truth audit found that labels are sampled from scenario rules, contradiction states are not independently represented, incidental generator missingness is only partly labelled, and several labels can be unobservable from normalized source rows. Accordingly, apparent false positives may include plausible omissions, while some false negatives reflect rule gaps. The reported score is therefore a strict measure of agreement with generator truth; independent semantic adjudication is treated as a separate clinical evaluation phase.

\subsection{Knowledge Graph and Retrieval}

The patient graph is stored as node and edge CSV files. Nodes represent the patient, scenario, laboratories, trends, missing context, questions, symptoms, medications, vital signs, lifestyle, family history, medical history, and wearables. Edges represent typed patient--evidence and evidence--question relations with source metadata. Evidence ranking is deterministic and combines scenario relevance, relationship relevance, contradiction, temporal relevance, detected-trend links, medication and symptom context, record quality, and provenance completeness. Retrieval is capped at the top 10 items per patient.

The graph is evaluated as provenance and evidence-lineage infrastructure. Current experiments measure loading, retrieval, source-row recoverability, orphan edges, and provenance completeness. Incremental graph effects on generated findings, omissions, unsupported claims, and reviewer preference are reserved for a dedicated paired graph/no-graph evaluation in which all other variables are held constant.

\subsection{Structured Synthesis, Operational Audit, and Paired Baseline}

Archived CareGraph synthesis uses the OpenAI API with Pydantic schema parsing. The model does not receive unrestricted raw tables in the CareGraph condition; it receives a patient-scoped deterministic insight object, ranked graph evidence, and the authorised registry $K_p$. Findings and next steps must cite authorised evidence keys, extra fields are forbidden, and AI and graph inputs can be required. The final analysed condition used model identifier \texttt{gpt-5.6}, simple reading level, prompt version \hashtt{caregraph-synthesis-v1.2}, prompt SHA-256 \hashtt{8a451b0426978c9a045cfcb17c93917d65f7d9d06bc29e5f70e152b9321cc27d}, and schema \hashtt{caregraph-synthesis-v1.1}. A deterministic fallback exists but was disabled in the graph-required engineering batch.

The pre-specified broad comparison protocol contained six conditions and three stochastic repetitions, but the complete 1{,}200-run design was not executed. Instead, a matched final-prompt cohort was formed from the 56 successful CareGraph outputs sharing the frozen final prompt hash. It contained 7 kidney-safety, 10 liver-enzyme, 10 anaemia/fatigue, 10 thyroid/endocrine, 10 medication-safety, and 9 inflammation/autoimmune cases; metabolic-risk and cardiovascular-risk cases had been generated under earlier prompt hashes and were excluded from the matched analysis.

For each of the 56 matched patients, an isolated monolithic baseline received only the normalised raw synthetic record with direct row-level source keys. Scenario labels, synthetic summaries, CareGraph trends, missing-context outputs, deterministic insights, recommendations, graph evidence, safety outputs, and temporal answer hints were excluded and recursively checked before the API call. The baseline used \texttt{gpt-5.6}, simple reading level, neutral prompt \hashtt{monolithic-baseline-v1.2-neutral} (SHA-256 \hashtt{97808536d0825ee5fec3b34b7ce40348ae9e3c8b721b5a87a93c73ad69efb382}), and a CareGraph-compatible schema capped at five findings, five next steps, and five clinician questions. This isolates architectural decomposition rather than model identity. One generation was run per patient, and the paired analysis compares pre-safety outputs because the monolithic condition was not passed through the CareGraph safety layer.

\subsection{Recommendation Safety Agent}

Ruleset \hashtt{recommendation-safety-v1.2} extends transparent patterns for explicit and implied diagnosis, medication initiation or cessation, dose adjustment, treatment planning, false reassurance, unsupported urgency, discouraging care, unsupported causality, and safe-question or uncertainty exceptions. Evidence-key checks remain separate from language-safety checks. The release score is:
\[
\text{Risk score} = 5 \times \text{hard violations} + \text{revision violations}.
\]

The held-out benchmark contains 39 examples: 20 unsafe and 19 safe. It was authored within the project and locked before v1.2 tuning, but it was not independently expert-adjudicated. Consequently, it supports a bounded rule-detection result rather than a comprehensive clinical-safety claim.

\section{Results}

\subsection{Held-Out Trend Classification}

The frozen OLS implementation evaluated once on 1{,}700 held-out labels achieved accuracy 0.827 (95\% CI 0.808--0.846), balanced accuracy 0.846, macro-F1 0.837 (95\% CI 0.819--0.854), and weighted F1 0.823. The explicit sufficiency gate achieved insufficient-data F1 0.974 (95\% CI 0.960--0.985).

\begin{table}[H]
\centering
\caption*{\textbf{Table 3.} Held-Out Trend Metrics by Class}
\label{tab:trend-metrics}
\small
\begin{tabular}{@{}lcccc@{}}
\toprule
\textbf{Class} & \textbf{Precision} & \textbf{Recall} & \textbf{F1} & \textbf{Support} \\
\midrule
Decreasing & 0.783 & 0.863 & 0.821 & 351 \\
Increasing & 0.826 & 0.870 & 0.847 & 469 \\
Insufficient data & 0.949 & 1.000 & 0.974 & 350 \\
Stable & 0.767 & 0.651 & 0.704 & 530 \\
\bottomrule
\end{tabular}
\end{table}

\begin{figure}[H]
\centering
\includegraphics[width=0.78\textwidth]{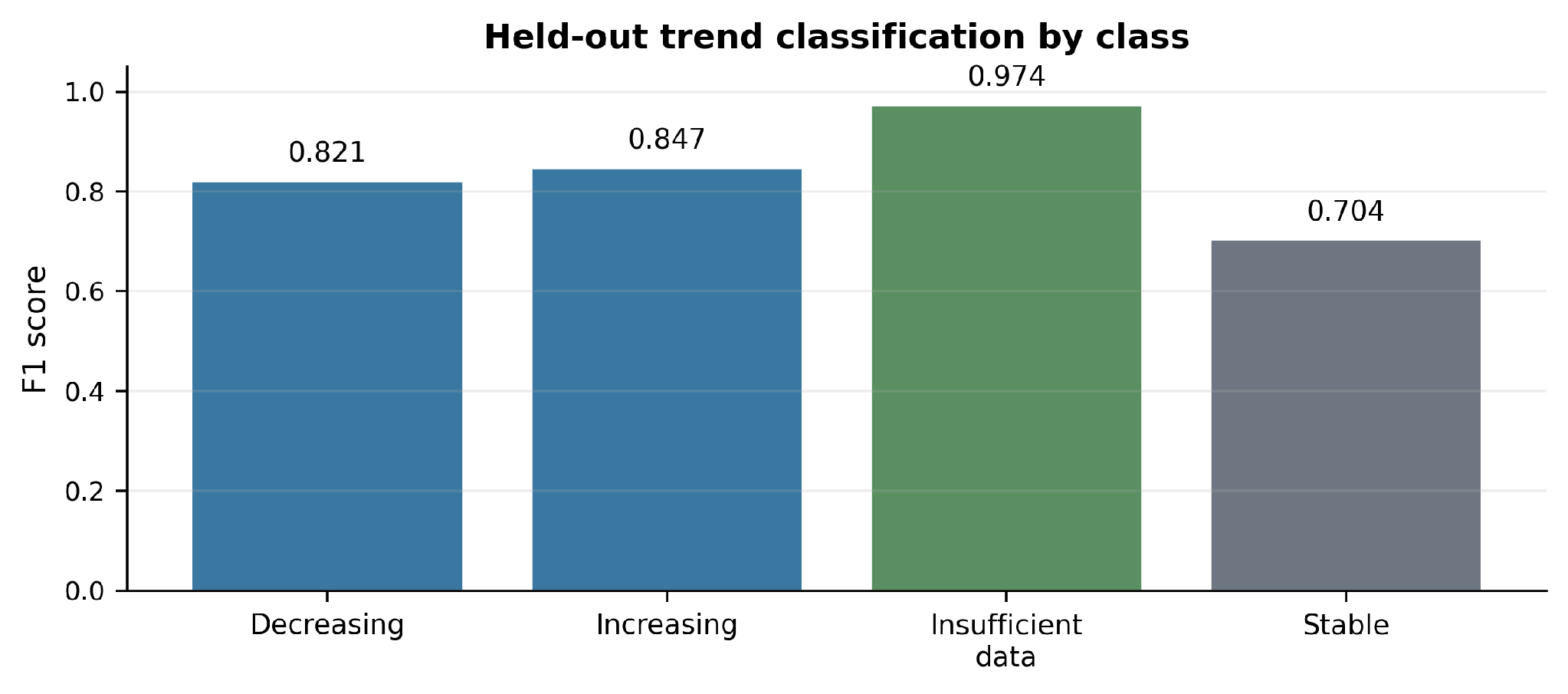}
\caption{Held-out trend F1 by class under the frozen OLS rule. The explicit sufficiency gate yields strong insufficient-data performance; stable classification remains the weakest class.}
\label{fig:trend-f1}
\end{figure}

The weakest scenario was inflammation/autoimmune signal (macro-F1 0.762), and ANA was the weakest biomarker (macro-F1 0.191). Validation macro-F1 was 0.846378 for OLS versus 0.846225 for endpoint, Theil--Sen, and ensemble methods; the OLS margin was only 0.000154. The principal correction was therefore the explicit sufficiency mechanism and frozen evaluation protocol, not a meaningful advantage of slope estimation.

\begin{figure}[H]
\centering
\includegraphics[width=0.78\textwidth]{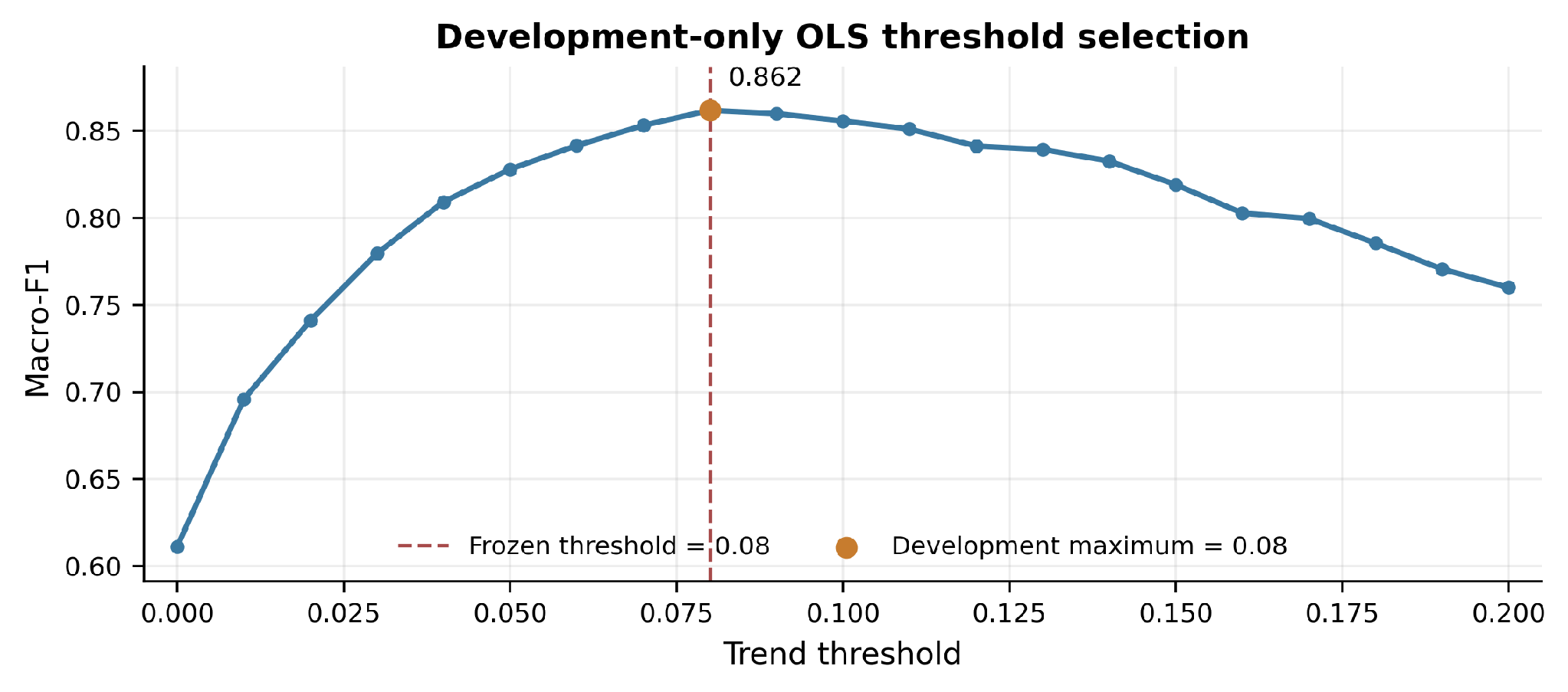}
\caption{Development-only threshold selection for OLS. The active threshold was frozen at 0.08 before replacement held-out evaluation.}
\label{fig:threshold}
\end{figure}

\subsection{Missing-Context Detection}

\begin{table}[H]
\centering
\caption*{\textbf{Table 4.} Held-Out Strict Missing-Context Results}
\label{tab:missing-context}
\small
\begin{tabular}{@{}lcccccc@{}}
\toprule
\textbf{Detector} & \textbf{TP} & \textbf{FP} & \textbf{FN} & \textbf{Precision} & \textbf{Recall} & \textbf{Micro-F1} \\
\midrule
Legacy & 177 & 371 & 387 & 0.323 & 0.314 & 0.318 \\
Version 2 & 542 & 224 & 22 & 0.708 & 0.961 & 0.815 \\
\bottomrule
\end{tabular}
\end{table}

Version 2 achieved strict precision 0.708, recall 0.961, micro-F1 0.815 (95\% CI 0.793--0.836), and macro-F1 0.874. The paired mean patient-level set-F1 difference from the legacy detector was 0.452 (95\% CI 0.406--0.496; Holm-adjusted $p = 5.89 \times 10^{-41}$; 400 pairs).

\begin{figure}[H]
\centering
\includegraphics[width=0.78\textwidth]{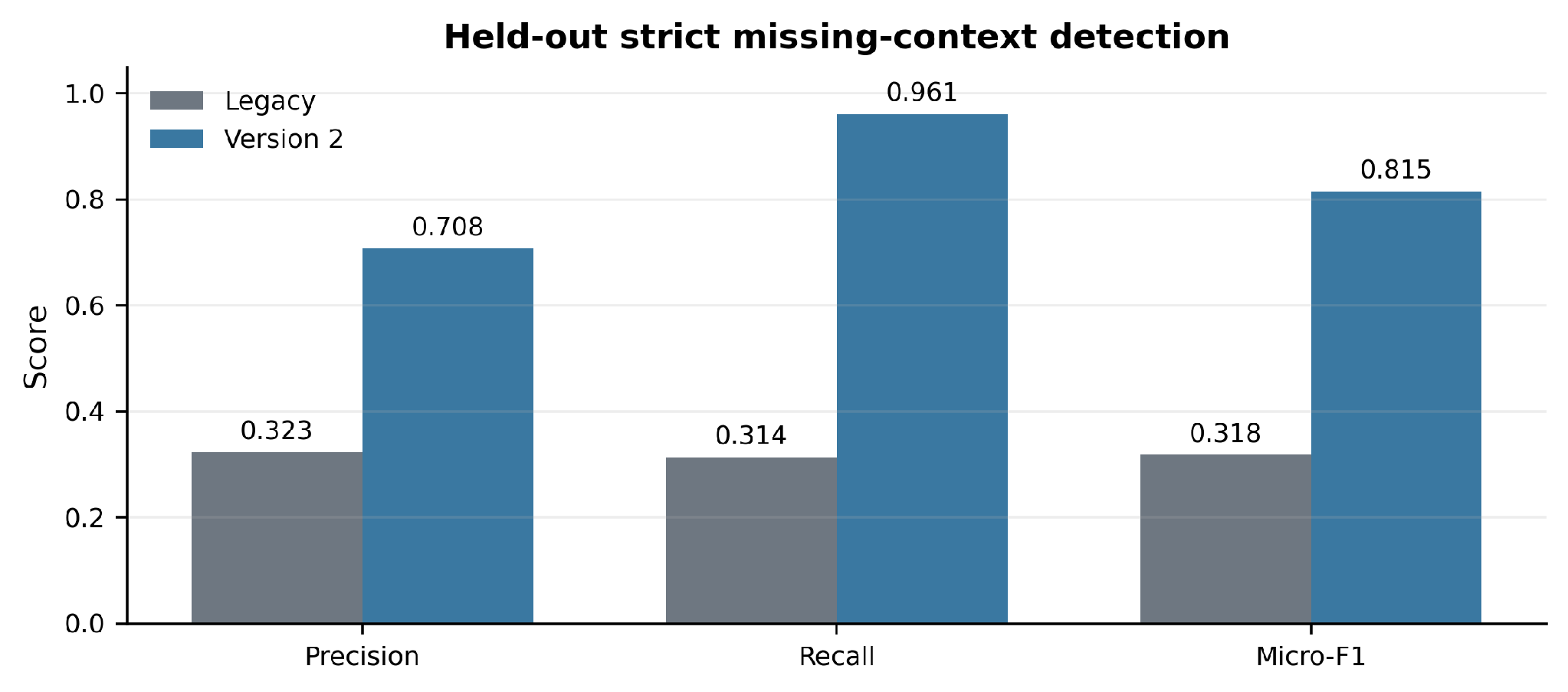}
\caption{Strict held-out missing-context performance. Version 2 substantially improves precision, recall, and micro-F1 over the legacy detector.}
\label{fig:missing-context}
\end{figure}

The export contains 224 benchmark false positives and 22 false negatives. The lowest scenario micro-F1 values were liver-enzyme trend (0.650) and medication safety/side-effect context (0.682); cardiovascular risk was highest at 0.924. Automated error labels are provisional, and no context-state confusion matrix or adjudicated semantic score is reported.

\subsection{Safety Rule Evaluation}

\begin{table}[H]
\centering
\caption*{\textbf{Table 5.} Held-Out Safety Benchmark}
\label{tab:safety}
\small
\begin{tabular}{@{}lccccccc@{}}
\toprule
\textbf{Ruleset} & \textbf{TP} & \textbf{TN} & \textbf{FP} & \textbf{FN} & \textbf{Precision} & \textbf{Recall} & \textbf{F1} \\
\midrule
v1.1 & 2 & 19 & 0 & 18 & 1.000 & 0.100 & 0.182 \\
v1.2 & 19 & 19 & 0 & 1 & 1.000 & 0.950 & 0.974 \\
\bottomrule
\end{tabular}
\end{table}

Ruleset v1.2 achieved precision 1.000, recall 0.950, F1 0.974, specificity 1.000, false-negative rate 0.050, and false-positive rate 0.000. Correctness increased by 0.436 relative to v1.1 (95\% CI 0.282--0.590; exact McNemar Holm-adjusted $p = 1.53 \times 10^{-5}$; 39 pairs). One treatment-planning paraphrase remained a false negative and is retained in the public error file.

\begin{figure}[H]
\centering
\includegraphics[width=0.78\textwidth]{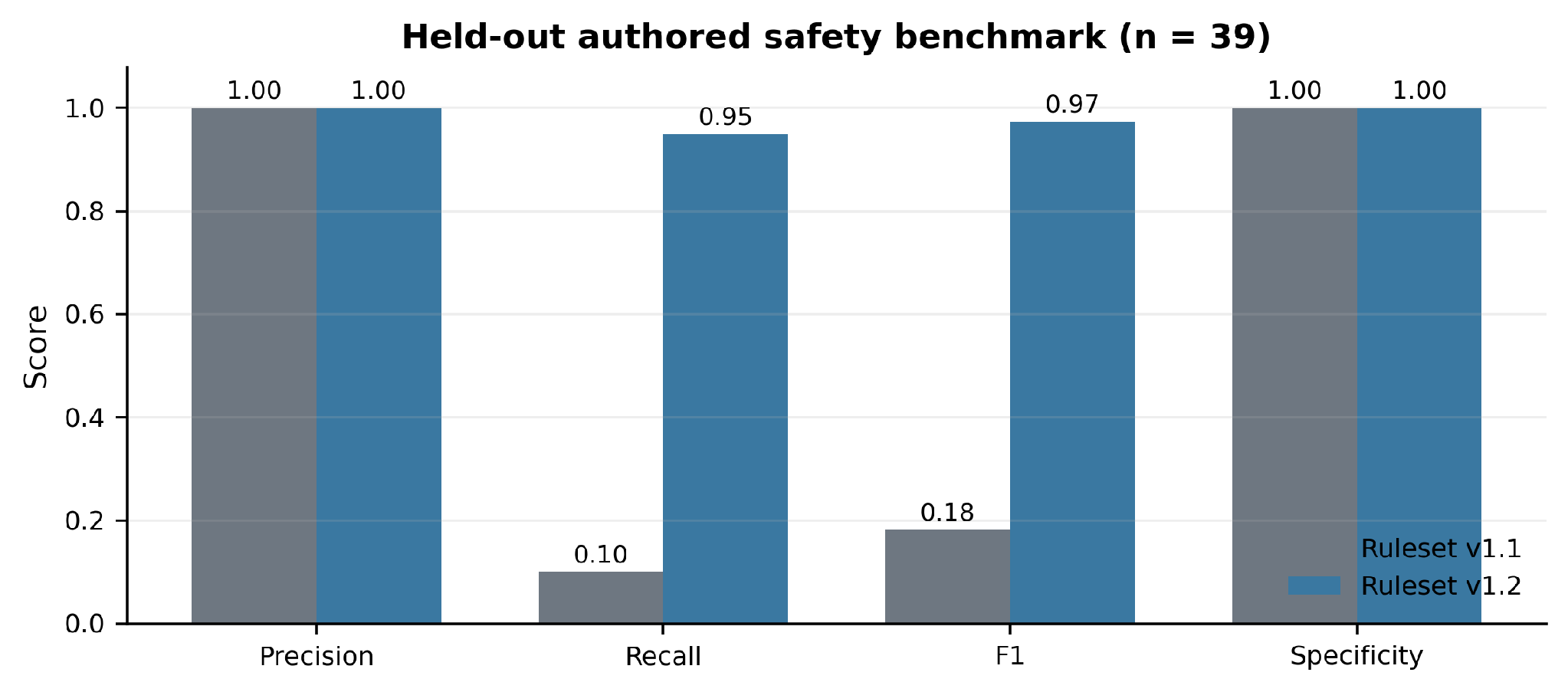}
\caption{Held-out authored safety benchmark. Ruleset v1.2 improves recall and F1 while retaining zero false positives in the 19 safe examples.}
\label{fig:safety}
\end{figure}

\subsection{Graph Retrieval and Provenance Audit}

\begin{table}[H]
\centering
\caption*{\textbf{Table 6.} Graph Structural and Provenance Results}
\label{tab:graph}
\small
\begin{tabular}{@{}lc@{}}
\toprule
\textbf{Measure} & \textbf{Value} \\
\midrule
Patients audited & 400 \\
Available graph-evidence items & 19{,}531 \\
Retrieved top-10 items & 4{,}000 \\
Mean nodes per patient & 46.0 \\
Mean saved edges per patient & 48.8 \\
Mean retrieval fraction & 0.207 \\
Orphan-edge rate & 0.000 \\
Source-row recoverability & 1.000 \\
Mean provenance-complete edge rate & 0.979 \\
\bottomrule
\end{tabular}
\end{table}

Retrieval and evidence-lineage mechanics were verified across the legacy 400-patient source snapshot. No graph citation rate, graph-specific finding rate, important-omission rate, unsupported-claim rate, reviewer preference, or content-quality effect is reported because paired graph/no-graph generation and review did not run.

\begin{figure}[H]
\centering
\includegraphics[width=0.78\textwidth]{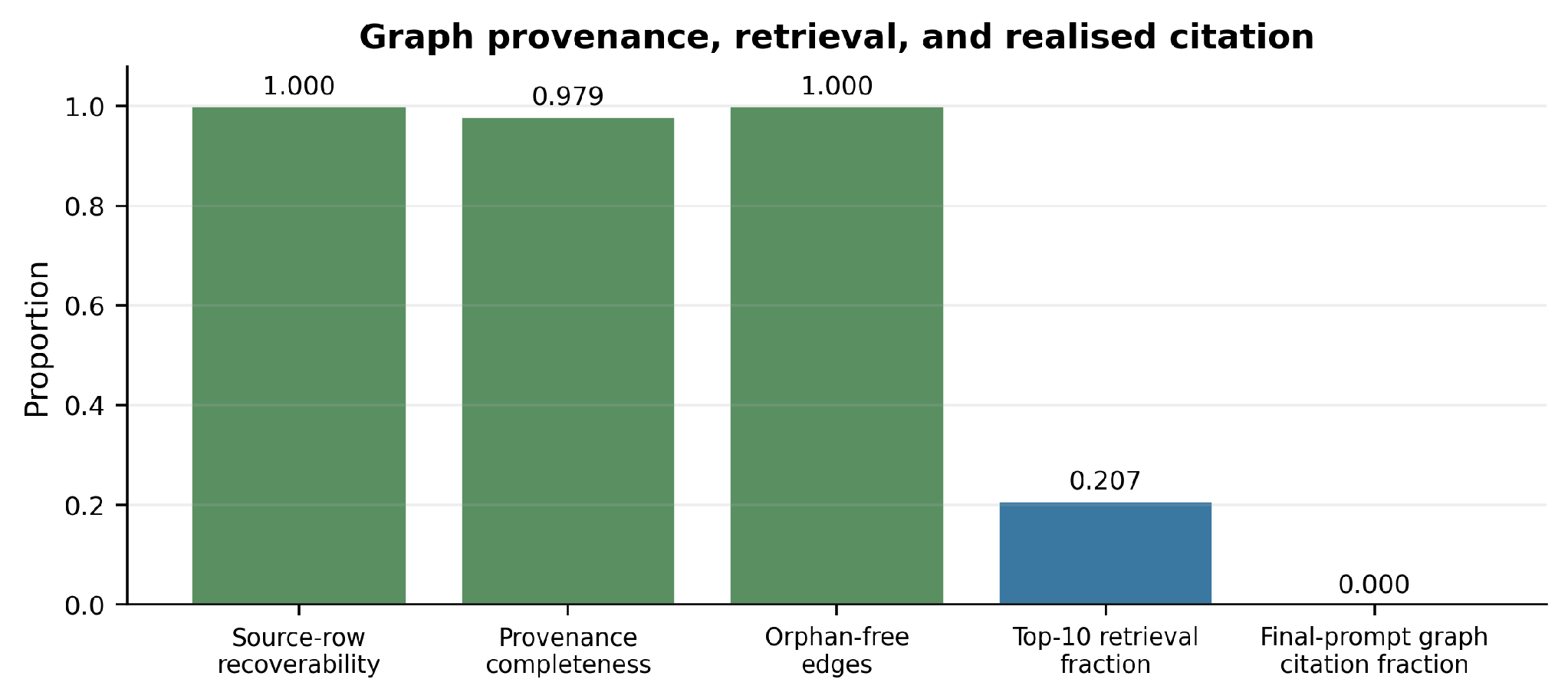}
\caption{Graph provenance, retrieval, and realised citation. Structural lineage metrics are high, but the final-prompt cohort cited no graph-specific evidence.}
\label{fig:graph}
\end{figure}

\subsection{80-Patient Operational Audit and Evidence Status}

\needspace{8\baselineskip}\begin{center}\small\textbf{Table 7.} Evaluation Coverage and Next-Phase Studies\end{center}\label{tab:coverage}
\vspace{-0.3em}
\begin{longtable}{@{}p{0.19\textwidth} p{0.15\textwidth} p{0.19\textwidth} p{0.27\textwidth}@{}}
\toprule
\textbf{Study element} & \textbf{Evaluated sample} & \textbf{Evidence in this study} & \textbf{Role in research program} \\
\midrule
\endfirsthead
\toprule
\textbf{Study element} & \textbf{Evaluated sample} & \textbf{Evidence in this study} & \textbf{Role in research program} \\
\midrule
\endhead
\bottomrule
\endfoot
Trend held-out test & 1{,}700 labels & Frozen one-shot evaluation & Core analytical validity \\[3pt]
Missing-context held-out test & 400 patients / 564 labels & Strict generator-truth evaluation & Core context-detection validity \\[3pt]
Safety held-out test & 39 examples & Frozen authored benchmark & Bounded safety-rule validity \\[3pt]
Graph provenance audit & 400 patients & Structural and lineage evaluation & Provenance infrastructure \\[3pt]
Condition C deterministic comparison & 80 patients & Schema and key-contract evaluation & Deterministic output contract \\[3pt]
80-patient graph-required AI batch & 80 scheduled & 79 syntheses; 78 presentations & Operational reliability and failure analysis \\[3pt]
Matched final-prompt comparison & 56 patient pairs & CareGraph versus isolated monolithic generation & System-level focus-versus-breadth analysis \\[3pt]
Monolithic leakage validation & 56 patients & Single neutral prompt/hash; all checks passed & Baseline isolation \\[3pt]
Graph/no-graph content study & Dedicated paired cohort & Next-phase causal evaluation & Isolate incremental graph contribution \\[3pt]
Matched safety study & Dedicated paired cohort & Next-phase release evaluation & Apply identical safety processing to both systems \\[3pt]
Blinded physician review & 56 pairs prepared & Clinical evaluation phase & Independent relevance, omission, safety, and preference scoring \\[3pt]
Prompt consistency observation & 79 successful syntheses & Sequential prompt evolution stratified by hash & Final-hash subset used for paired comparison \\
\end{longtable}

All 80 deterministic condition-C outputs passed schema validation, existing-key evidence validation, evidence completeness, and layered provenance checks. These results establish the structural contract of the deterministic pathway. Clinical relevance, prioritization, omissions, and preference are distinct questions addressed by the matched generation study and the next clinical-adjudication phase.

The separate full-pipeline audit scheduled 80 patients, with 10 from each scenario. AI synthesis completed for 79 patients (98.75\%; Wilson 95\% CI 93.3\%--99.8\%), the graph loaded for all 79 successful syntheses, no fallback was used, and presentation completed for 78 patients (97.5\%; Wilson 95\% CI 91.3\%--99.3\%). One synthesis was blocked before presentation and one run failed at the evidence-key validator. Across the 79 archived successful syntheses, no invalid source keys remained, and no priority finding or next step lacked an evidence key.

The 79 successful syntheses averaged 37.45~s for the AI stage (median 37.99~s; 95th percentile 49.75~s), 9{,}764 input tokens, 2{,}904 output tokens, and 12{,}668 total tokens. They produced a mean of 4.30 findings, 4.05 next steps, 4.61 clinician questions, and 15.33 evidence references. Registry utilisation was 1{,}211 referenced keys out of 2{,}291 available keys (52.9\%). Graph citation utilisation was 42 of 790 available graph-evidence items (5.3\%), with graph citations in 11 of 79 patients; all graph citations occurred under the earliest prompt condition.

The batch was not prompt-homogeneous. Three prompt hashes occurred under the same \hashtt{caregraph-synthesis-v1.2} label: 19, 4, and 56 successful patients, respectively. The latest 56-patient condition averaged 40.15~s and 13{,}176 total tokens per synthesis, compared with 27.14~s and 10{,}989 tokens in the earliest 19-patient condition. The latest condition cited no graph-evidence keys. These differences are descriptive because prompt changes were introduced sequentially rather than randomised.

Safety review returned needs\_revision for 78 syntheses and blocked for one. The rules appended standard diagnosis, scope, and medication limitations, but operational review also exposed a medication-change pattern that matched benign phrases such as ``start date,'' ``weight change,'' and laboratory ``decrease'' near medication context. Fifty-six findings or steps were removed across 44 patients. Text inspection indicates that these removals were dominated by apparent false positives, including the blocked CG-0308 record-quality suggestion to confirm medication dose, frequency, and start date. Within the final-prompt matched subset, 55 of 56 outputs were marked safe for presentation and one was blocked; 40 items were removed across 33 patients. This operational finding is retained as a safety-generalisation limitation rather than combined with the small authored benchmark.

CG-0354 failed closed because the model emitted \texttt{insight:3:insufficient\_longitud\_data}, which was not an authorised source key. With \texttt{-{}-require-ai} enabled, the validator stopped the pipeline before safety review or presentation. This failure demonstrates effective release gating but motivates a bounded validation retry or dynamically constrained evidence-key enumeration.

\subsection{Matched 56-Patient Monolithic Comparison}

\begin{figure}[H]
\centering
\includegraphics[width=\textwidth]{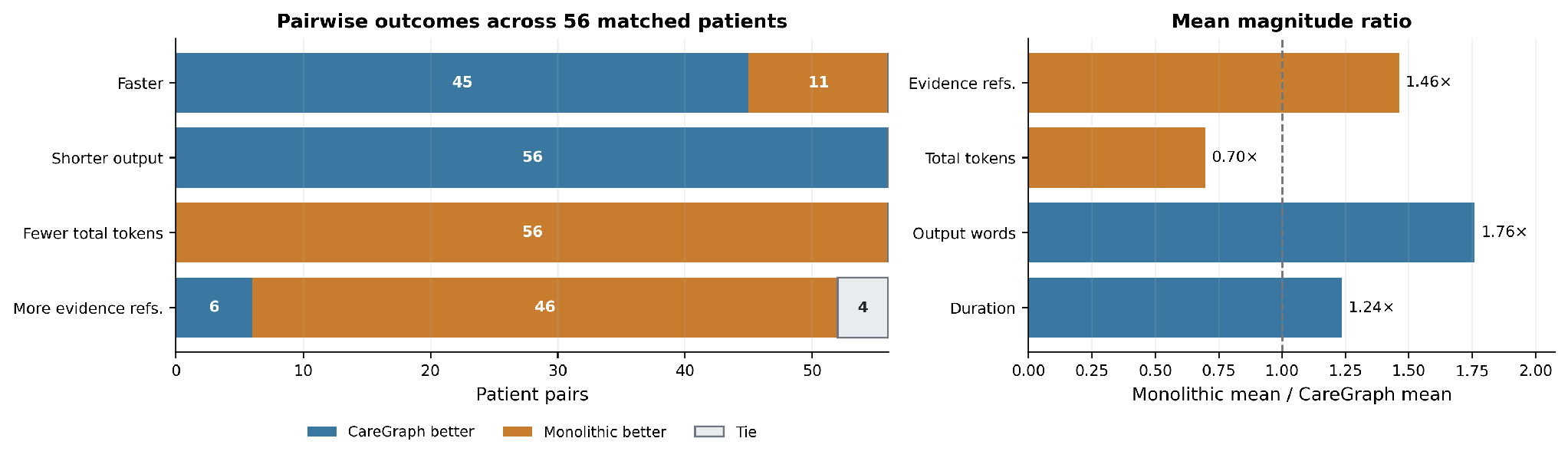}
\caption{Matched generation trade-offs across 56 patients. CareGraph wins most latency pairs and every output-length pair; the monolithic baseline wins every total-token pair and most evidence-reference pairs. Mean ratios use CareGraph as the denominator.}
\label{fig:comparison}
\end{figure}

All 56 monolithic runs completed under one neutral prompt hash, one model, one schema, and one reading level; input-leakage validation passed for every patient. All 56 matched CareGraph outputs shared the final prompt hash. CareGraph generation was faster in 45 of 56 pairs and produced fewer words in all 56 pairs. The monolithic baseline used fewer total tokens in every pair and cited more evidence references in 46 pairs, with four ties. Fig.~\ref{fig:comparison} visualises these paired trade-offs.

Table 8 reports paired generation metrics. CareGraph was 9.46~s faster on average and produced 502 fewer words, but used 3{,}994 more total tokens because its structured upstream evidence payload was substantially larger. The monolithic baseline produced 0.45 more findings and 6.80 more evidence references on average. Mean evidence-registry utilisation did not differ significantly after Holm correction (0.534 versus 0.504; adjusted $p = 0.171$). These quantities measure generation behaviour and traceability breadth, not clinical correctness.

\begin{table}[H]
\centering
\caption*{\textbf{Table 8.} Matched 56-Patient Generation Comparison}
\label{tab:comparison}
\small
\begin{tabular}{@{}lccccc@{}}
\toprule
\textbf{Metric} & \textbf{Monolithic mean} & \textbf{CareGraph mean} & \textbf{Mean diff.\ M$-$C} & \textbf{Holm $p$} & \textbf{Rank-biserial} \\
\midrule
Duration (s) & 49.62 & 40.15 & +9.46 & $2.74\times10^{-7}$ & 0.821 \\
Input tokens & 5{,}172 & 10{,}143 & $-$4{,}971 & $6.79\times10^{-10}$ & $-$1.000 \\
Output tokens & 4{,}010 & 3{,}033 & +976 & $6.63\times10^{-9}$ & 0.936 \\
Total tokens & 9{,}182 & 13{,}176 & $-$3{,}994 & $6.79\times10^{-10}$ & $-$1.000 \\
Output words & 1{,}163 & 661 & +502 & $6.79\times10^{-10}$ & 1.000 \\
Findings & 4.86 & 4.41 & +0.45 & $1.37\times10^{-3}$ & 0.794 \\
Evidence references & 21.52 & 14.71 & +6.80 & $2.66\times10^{-8}$ & 0.930 \\
Evidence utilisation & 0.534 & 0.504 & +0.030 & 0.171 & 0.211 \\
\bottomrule
\end{tabular}
\end{table}

Exploratory lexical checks compared each output with predefined deterministic targets. CareGraph mentioned the expected trend direction in 229 of 244 trend records (93.9\%), versus 131 of 244 (53.7\%) for the monolithic baseline. It mentioned expected missing-context terms in 96 of 99 records (97.0\%), versus 67 of 99 (67.7\%). These checks are sensitive to wording and benchmark design and are not substitutes for clinician adjudication. The monolithic outputs were consistently broader and longer, whereas CareGraph was more focused on the upstream longitudinal targets.

No final-prompt CareGraph output cited a graph-evidence key, despite graph loading and graph-evidence availability for every matched patient. The paired study therefore attributes the observed target alignment to the overall evidence-governed architecture rather than to a demonstrated incremental graph effect. Automated coverage checks characterize benchmark alignment, while expert clinical adjudication is treated as a separate evaluation phase and is not included in the present engineering results.

\section{Discussion}

\subsection{CareGraph as an Extensible Health-Intelligence Backbone}

CareGraph's principal contribution is the architecture that sits between fragmented health data and the final explanation. The system creates a stable patient state and authorized evidence registry into which multiple data sources can feed. Deterministic agents calculate and qualify longitudinal patterns; the graph preserves relationships and provenance; the LLM organizes only approved evidence; and safety and release gates control what can be presented. This separation makes the framework extensible: new sources can be added through adapters without turning the entire pipeline into an opaque generative prompt. The controlled experiments in this paper evaluate that backbone at component, operational, and comparative levels.

\subsection{Controlled Evidence for Core Components}

The held-out component results show that transparent mechanisms can provide a strong analytical foundation for personalized synthesis. The explicit data-sufficiency gate was more consequential than the small difference between endpoint and slope-based trend candidates, producing insufficient-data F1 of 0.974. Schema alignment substantially improved missing-context detection from micro-F1 0.318 to 0.815, although precision of 0.708 and scenario-conditioned truth indicate that strict benchmark success is not equivalent to complete clinical relevance. These findings support a modular design in which each analytical contract can be tested and improved independently.

\subsection{Provenance Infrastructure and Safety Governance}

The graph layer currently contributes a durable relationship and provenance substrate rather than a demonstrated graph-reasoning advantage. The 400-patient audit verified zero orphan edges, complete source-row recoverability, and mean provenance completeness of 0.979. Low explicit graph citation in the generation cohort indicates that deterministic summaries often supplied the most concise evidence for the LLM. This is an architectural finding, not a failure of the graph: the present study verifies lineage and retrieval, while a controlled graph/no-graph study will isolate incremental content value. The safety layer likewise demonstrated the value of an independent release boundary by blocking or revising output, while the batch analysis exposed surface-pattern false positives that can be corrected without redesigning the core system.

\subsection{Focus Versus Breadth in the Matched Baseline}

The matched 56-patient comparison clarifies the design trade-off. CareGraph produced faster and markedly shorter outputs that were more consistently aligned with predefined longitudinal trends and missing-context targets. The monolithic baseline used fewer total tokens, cited more direct raw-record evidence, and surfaced broader secondary context. This distinction is important: CareGraph is not intended to maximize the number of observations mentioned. It is intended to convert fragmented longitudinal evidence into a focused, traceable intelligence layer with explicit failure and release states. The monolithic condition remains a useful breadth-oriented comparator.

\subsection{Research Implications}

Within the predefined synthetic longitudinal-synthesis objective, the supported system-level conclusion is that CareGraph is the better-aligned architecture. Its advantage arises from disciplined decomposition: patient-specific evidence is calculated before language generation, every required finding and action must cite an authorized key, and invalid or non-presentable outputs fail closed. The result is not a claim that CareGraph contains more universal medical knowledge than a monolithic LLM. It is evidence that a modular hybrid AI backbone can produce more focused and governable personalized health intelligence while retaining an inspectable path back to the underlying record. This is the central research contribution and the basis for extending CareGraph to broader data sources and future clinical evaluation.

\section{Study Scope and Threats to Validity}

\needspace{8\baselineskip}\begin{center}\small\textbf{Table 9.} Principal Threats to Validity\end{center}\label{tab:threats}
\vspace{-0.3em}
\begin{longtable}{@{}p{0.24\textwidth} p{0.68\textwidth}@{}}
\toprule
\textbf{Category} & \textbf{Threat} \\
\midrule
\endfirsthead
\toprule
\textbf{Category} & \textbf{Threat} \\
\midrule
\endhead
\bottomrule
\endfoot
Synthetic coupling & Generator scenarios and missingness may favour schema-aligned rules and do not reproduce clinical prevalence, ambiguity, or documentation practice. \\[4pt]
Trend construct & The scenario-to-expected-lab inventory remains coupled to the generator schema; stable and selected biomarker results are weaker than directional classes. \\[4pt]
Missing-context truth & Strict labels are scenario-conditioned and omit some plausible or contradictory states; current metrics measure agreement with generator truth rather than complete clinical relevance. \\[4pt]
Safety benchmark & Only 88 authored examples were available across development and held-out sets; the benchmark has no independent expert annotation and retains one held-out treatment-plan miss. \\[4pt]
Operational safety specificity & Medication-change rules matched benign documentation phrases and cross-sentence proximity; generated-output specificity requires sentence-bounded refinement. \\[4pt]
Matched comparison cohort & The 56-patient final-prompt cohort represents six scenario groups and is not balanced across all eight; paired results describe this matched subset. \\[4pt]
Comparison processing & CareGraph and monolithic inputs differ by design, outputs were generated once per patient, and identical post-generation safety processing was not part of the present comparison. \\[4pt]
Clinical evaluation phase & Clinical relevance, important omissions, prioritization, usefulness, and preference were not part of the current engineering evaluation; a blinded physician-review packet has been prepared for a separate phase. \\[4pt]
Graph contribution & Current work verifies provenance and deterministic retrieval; incremental content impact requires a paired graph/no-graph study. \\[4pt]
Evidence semantics & Existing-key validity does not prove that generated language semantically follows the cited source. \\[4pt]
Prompt evolution & The 80-patient engineering batch used three prompt hashes sequentially under one version label; the final-hash subset was isolated for paired analysis. \\[4pt]
Controlled synthetic scope & No real clinical workflow or patient-outcome evaluation was performed; claims are limited to controlled synthetic performance and system behavior. \\
\end{longtable}

\section{Ethical, Privacy, and Safety Considerations}

The evaluated component cohorts are synthetic, and no identified real-patient data were used. Synthetic data reduce direct privacy risk but can encode unrealistic assumptions, demographic simplifications, and hidden fairness failures. Future work with real data would require an institutional determination, explicit intended use, access controls, encryption, audit logging, retention and incident-response policies, and a jurisdiction-specific regulatory assessment \citep{who2021, fda2022}.

CareGraph is intended for health-information organisation, longitudinal explanation, missing-context identification, and preparation for clinician discussion. It does not establish diagnosis, prescribe treatment, alter medication, provide emergency triage, or replace professional judgment. The safety rules reduce a bounded set of surface-language risks but are not a safety guarantee.

FHIR and controlled terminologies may improve interoperability, but standard mapping must be evaluated rather than assumed \citep{hl7fhir2023}. Untrusted documents also introduce prompt-injection, parser, unit-corruption, and fabricated-record risks; extraction, validation, synthesis, and presentation should remain separately governed.

\section{Translation Roadmap and Next-Phase Evaluation}

\begin{itemize}[leftmargin=1.4em]
\item \textbf{Reliability and release engineering.} The immediate systems phase is to freeze one uniquely versioned synthesis prompt and corrected safety rules, repeat the end-to-end cohort under immutable settings, and archive response identifiers, runtime, token use, costs, hashes, and failure metadata. This will estimate full-cohort reliability without sequential prompt drift.
\item \textbf{Clinical evaluation.} The prepared blinded physician-review study should independently score relevance, important omissions, factual support, prioritization, safety, usefulness, and pairwise preference. A second clinician on a stratified subset would permit inter-rater agreement and convert the current engineering comparison into a clinically adjudicated evaluation.
\item \textbf{Graph contribution and semantic evidence.} A paired graph/no-graph generation study with identical patient state, prompt, model, schema, limits, repetitions, and safety processing should isolate the graph's incremental value. Independent semantic review should additionally distinguish valid key membership from faithful use of the cited evidence.
\item \textbf{Interoperability and data expansion.} The backbone should be extended through governed adapters for FHIR-based EHR data, laboratory portals, medication histories, patient-reported outcomes, and wearable platforms. New sources should enter through the same patient-state, provenance, and evidence-registry contracts rather than bypassing them with unrestricted prompting.
\item \textbf{Prospective translation and reproducibility.} After external data governance, security, and intended-use review, CareGraph should be evaluated in real workflows for comprehension, trust, clinician-discussion quality, and downstream action. The public codebase should be accompanied by a tagged release, software and data licenses, frozen manifests, and an archival snapshot so that each reported result maps to an immutable experimental state.
\end{itemize}

\section{Conclusion}

CareGraph was evaluated as an auditable hybrid AI framework for personalized longitudinal health intelligence rather than as a single generative model. Under independently seeded synthetic evaluation, the frozen trend implementation achieved held-out accuracy 0.827 and macro-F1 0.837, including insufficient-data F1 0.974; the schema-aligned missing-context detector achieved micro-F1 0.815 versus 0.318 for the legacy detector; and the revised safety rules achieved held-out F1 0.974 on a small authored benchmark. The graph audit verified source-row recoverability, provenance completeness, orphan-free edges, and deterministic bounded retrieval.

The 80-patient graph-required stress test demonstrated operational completion and failure-aware governance: 79 AI syntheses completed, 78 reached presentation, no fallback was used, one output was blocked, and one invalid evidence key stopped the pipeline before release. In the matched 56-patient final-prompt comparison, CareGraph was faster, markedly shorter, and more aligned with predefined longitudinal targets, while the isolated monolithic baseline was broader, more token-efficient, and cited more direct raw-record evidence. These findings establish a clear system-level trade-off: CareGraph prioritizes focused, traceable, and release-governed intelligence, whereas the monolithic baseline prioritizes breadth.

CareGraph therefore establishes the backbone of a personalized health-intelligence system. Heterogeneous longitudinal sources can be normalized into a shared patient state, transformed into versioned evidence, organized through explicit provenance, and synthesized into an understandable account of what changed, what context is missing or conflicting, and what questions merit discussion. Within this controlled study, CareGraph is the stronger system for evidence-governed longitudinal synthesis because it combines analytical focus with inspectable evidence and fail-closed release boundaries. Its architecture is designed to expand through new data adapters and clinical evaluation rather than through a redesign of the core pipeline. CareGraph remains a research framework for health-information organization and clinician-discussion preparation---not a diagnostic model, treatment recommender, autonomous clinical decision-maker, or clinically validated medical device---but it provides a serious and technically grounded path toward patient-facing and clinician-facing AI personalized health intelligence.

\section*{Acknowledgment}
No acknowledgments apply.

\section*{Funding}
This research received no external funding.

\section*{Conflict of Interest}
The authors declare no competing interests.

\section*{Ethics Statement}
The reported study used entirely synthetic records and did not involve human participants or identifiable private information. Formal institutional review was not obtained because no human-subject data or clinician ratings are included in the present analysis.

\section*{Data and Code Availability}
Public code repository: {\small\url{https://github.com/PratikGhawate/ai-personalized-health-intelligence}}. The repository and accompanying submission materials should identify the evaluated state through a release tag or commit, software and data licenses, frozen prompt and configuration hashes, cohort manifests, component metrics, operational batch logs, the 56-patient monolithic bundle, and paired-analysis files. An archival snapshot or DOI can be added when the submission release is frozen.

\end{document}